\documentclass{article}
\usepackage{INTERSPEECH_v2,amsmath,graphicx,algpseudocode,algorithm,bm,cite,setspace,overpic}
\usepackage{lipsum}

\title{Adversarial Auto-encoders for Speech Based Emotion Recognition}
\name{Saurabh Sahu$^1$, Rahul Gupta$^2$, Ganesh Sivaraman$^1$, Wael AbdAlmageed$^3$, Carol Espy-Wilson$^1$}
\address{$^1$Speech Communication Laboratory, University of Maryland, College Park, MD, USA \\
$^2$Amazon.com, USA\\
$^3$VoiceVibes, Marriottsville, MD; Information Sciences Institute, USC, Los Angeles, CA, USA 
}  
\email{\{ssahu89,ganesa90,espy\}@umd.edu, gupra@amazon.com, wamageed@gmail.com}

\begin{document}
\ninept
\maketitle
%

\begin{abstract}
Recently, generative adversarial networks and adversarial auto-encoders have gained a lot of attention in machine learning community due to their exceptional performance in tasks such as digit classification and face recognition.
They map the auto-encoder's bottleneck layer output (termed as code vectors) to different noise Probability Distribution Functions (PDFs), that can be further regularized to cluster based on class information.
In addition, they also allow a generation of synthetic samples by sampling the code vectors from the mapped PDFs.
Inspired by these properties, we investigate the application of adversarial auto-encoders to the domain of emotion recognition.
Specifically, we conduct experiments on the following two aspects: (i) their ability to encode high dimensional feature vector representations for emotional utterances into a compressed space (with a minimal loss of emotion class discriminability in the compressed space), and (ii) their ability to regenerate synthetic samples in the original feature space, to be later used for purposes such as training emotion recognition classifiers. 
We demonstrate promise of adversarial auto-encoders with regards to these aspects on the Interactive Emotional Dyadic Motion Capture (IEMOCAP) corpus and present our analysis. 

\end{abstract}

\section{Introduction}
\label{sec:intro}

Emotion recognition has implications in psychiatry \cite{tacconi2008activity}, medicine \cite{maier2011emotion}, psychology \cite{JCLP:JCLP2270420302} and design of human-machine interaction systems \cite{cowie2001emotion}. Several research studies have focused on the prediction of the emotional state of a person based on cues from their speech, facial expression as well as physiological signals \cite{busso2004analysis, koelstra2012deap}.
The design of these systems typically requires extraction of a considerably large dimensionality of features to reliably capture the emotional traits, followed by training of a machine learning system.
However, this design inherently suffers from the following two drawbacks: (1) it is difficult to analyze the feature representations for the utterances due to the high dimensionality of the features used to represent the said utterances, and (2) obtaining a large dataset for training machine learning models on behavioral data is often restricted as the data collection is expensive and prohibitive in time. 
We address these issues in this paper by testing the recently proposed adversarial auto-encoder \cite{makhzani2015adversarial} setup for emotion recognition.
Auto-encoders have been known to learn compact representations for large dimensional subspaces \cite{baldi2012autoencoders}.
Adversarial auto-encoders further this concept by enforcing the auto-encoder codes to follow an arbitrary prior distribution, which can also be optionally regularized to carry class information. 
The goal of this paper is to investigate the application of auto-encoders to enhance the state of research in emotion recognition.

Emotion recognition is a fairly widely researched topic. 
Some of the previous works include use of F0 contours \cite{williams1972emotions, banziger2005role}, formant features, energy related features, timing features, articulation features, TEO features, voice quality features and spectral features for emotion recognition \cite{el2011survey}. 
Researchers have also investigated various machine learning algorithms such as Hidden Markov Models \cite{lin2005speech}, Gaussian Mixture Models (GMM) \cite{hu2007gmm}, Artificial Neural Networks \cite{singh2013ann}, Support Vector Machines (SVM)\cite{ ververidis2006emotional} and binary decision trees \cite{lee2011emotion} for emotion classification. 
Recently, researchers have also proposed several deep learning approaches for emotion recognition \cite{huangchei}.
Stuhlsatz et al. \cite{stuhlsatz2011deep} reported accuracies using a Deep Neural Network on 9 corpora using Generalized Discriminant Analysis features to do a binary classification between positive and negative arousal and positive and negative valence states. 
Xia and Liu \cite{68cb78fa64ae4f0d837c08a414b189f2} implemented a denoising auto-encoder for emotion recognition. 
They captured the neutral and emotional information by mapping the input to two hidden representations, and later using an SVM model for further classification. 
Ghosh et al. \cite{ghosh2015learning} used denoising auto-encoders and showed that the bottleneck layer representations are highly discriminative of activation intensity and at distinguishing negative versus positive valence. 
A typical setup in several of these studies involves using a large dimensionality of features and using a machine learning algorithm to learn class boundaries in the corresponding feature space.
This design renders a joint feature analysis in the high dimensional space rather difficult. 
Adversarial auto-encoders address these issues by encoding a high dimensional feature vector onto a code vector, which can be further enforced to follow an arbitrary Probability Distribution Function (PDF).
They have been shown to perform quite well in digit recognition tasks and face recognition \cite{makhzani2015adversarial}.
We use adversarial auto-encoders for emotion recognition in this paper motivated by their performances on other tasks for feature compression as well as data generation from random noise samples.
To the best of our knowledge, this is the first such application of adversarial auto-encoders to the domain of emotion recognition. 

We borrow a specific setup of adversarial auto-encoders with adversarial regularization to incorporate class label information \cite{makhzani2015adversarial}.
After training the adversarial auto-encoders on utterances with emotions, we conduct two specific experiments: (i) classification using auto-encoders code vectors (as output by the adversarial auto-encoder's bottleneck layer) to investigate the discriminative power retained by the low dimensional features and, (ii) classification using a set of synthetically generated samples from the adversarial auto-encoder. 
We initially provide a background of the adversarial auto-encoders in the next section, followed by a detailed explanation of the application of adversarial auto-encoders for emotion recognition in Section~\ref{sec:adv_emo}. 
This section also provides detail of the dataset used in our experiments, as well as details on the two classification experiments.
The first classification experiment investigates the discriminative power retained by various dimensionality  reduction techniques (adversarial auto-encoder, auto-encoder, Principle Component Analysis, Linear Discriminant Analysis) in a low dimensional subspace. 
The second classification experiment investigates the use of generated synthetic vectors in training an emotion recognition classifier under two settings: (i) using synthetic data only, and (ii) appending synthetic data to the real dataset. 
We finally present our conclusions in Section~\ref{sec:conclusion}.

\section{Background on adversarial auto-encoders}
\begin{figure}[t]
\centering
\includegraphics[trim=0cm 6cm 16cm 0cm,clip=true,scale=.4]{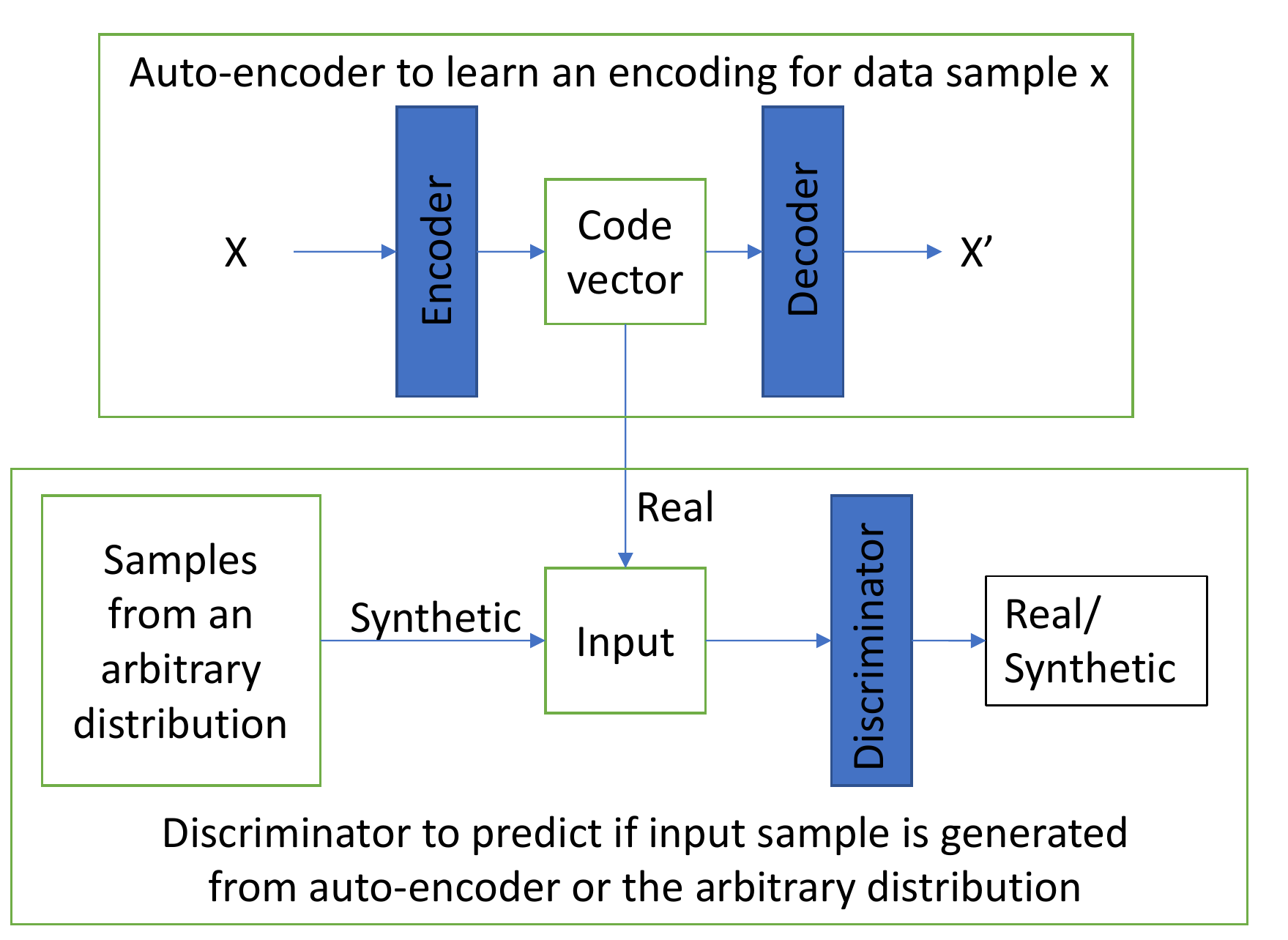}
\captionsetup{justification=centering}
\caption{A summarization of the adversarial auto-encoders.
The generator at the top creates code vectors. The discriminator learns to classify the code vectors generated from real data from the synthetic samples.}
\label{fig:adv_auto}
\vspace{-4mm}
\end{figure}

Makhzani et al. \cite{makhzani2015adversarial} proposed adversarial auto-encoders based on Generative Adversarial Networks \cite{goodfellow2014generative} consisting of a generator and a discriminator. 
Figure~\ref{fig:adv_auto} summarizes the framework for the adversarial auto-encoders.
An adversarial auto-encoder broadly consists of two major components: a generator and a discriminator. 
In Figure ~\ref{fig:adv_auto}, we show the generator at the top, which given a sample $x$ from the real data (e.g. pixels from an image, features from a speech sample) learns a code vector for the data sample. We model an auto-encoder for this purpose, where the model learns to reconstruct $x$ through a bottleneck layer. 
We represent the reconstruction for $x$ as $x^\prime$ in Figure~\ref{fig:adv_auto}. 
The discriminator (in the bottom half of Figure~\ref{fig:adv_auto}) obtains the code vectors encoded by the auto-encoder as well as synthetic samples from an arbitrary distribution, and learns to discriminate the real samples from the synthetic sample. 
The generator and the discriminator operate against each other, where the discriminator attempts to accurately classify real samples against synthetic samples and the generator produces code vectors to confuse the discriminator (so that the discriminator is not able to distinguish real from synthetic inputs). 
Makhzani et al. \cite{makhzani2015adversarial} performed extensive set of experiments to demonstrate the utility of adversarial auto-encoders.
They further proposed tricks such as, in a setting where the samples $x$ belong to different classes, the arbitrary distribution is a mixture of Probability Distribution Functions (PDF) with as many components as the number of classes.
Furthermore, to enforce each component of the mixture PDF to correspond to a class, the authors regularized the hidden code vector generation by providing a one-hot encoding for the classes to the discriminator (Figure 3 in \cite{makhzani2015adversarial}).
We refer the reader to \cite{makhzani2015adversarial} for further details regarding the optimization of adversarial auto-encoders and given this background for the adversarial auto-encoders, we motivate their use for emotion recognition.

\section{Adversarial auto-encoders for emotion recognition}
\label{sec:adv_emo}
Emotion recognition from speech is a classical problem and a typical setup for emotion recognition involves training a machine learning model (e.g. a classifier, regressor) on a set of extracted features.
However, in order to maximally capture the difference between the emotion classes, these models often need a high dimensionality of features. 
Despite promising performances reported with such an approach, the analysis of data-points on a high dimensional feature space is challenging.  
Furthermore, given that these samples are often obtained from curated datasets and then annotated for emotional content, the dataset size is limited. 
We address these issues using adversarial auto-encoders. 
Specifically, our experiments are geared towards investigating the following two aspects on: (i) compressing the high dimensional feature vectors to a small dimensionality with minimal loss in their discriminative power and, (ii) generating synthetic samples using the adversarial auto-encoders to address the data sparsity issue typically associated with this domain.  

\subsection{Dataset}
We used the Interactive Emotional Dyadic Motion Capture (IEMOCAP) dataset for our experiments \cite{busso2008iemocap}. 
It comprises of a set of scripted and spontaneous dyadic interactions sessions performed by actors. 
There are 5 such sessions with two actors each (one female and one male) and each session has different actors participating in them. 
The dataset consists of approximately 12 hours of speech from 10 human subjects. 
The interactions have been segmented into utterances each 2-5 seconds long which are then labeled by three annotators for emotion labels such as happy, sad, angry, excitement, neutral, and frustration. 
For our classification experiments we only focused on a set of 4490 utterances shared amongst four emotional labels: neutral (1708), angry (1103), sad (1084), and happy (595).
These utterances have a majority agreement amongst the annotators (at least 2/3 annotators) regarding the emotion label.  

\subsection{Features}
We extract a set of 1582 features using the openSMILE toolkit \cite{eyben2013recent}.
The set consists of an assembly of spectral, prosody and energy based features.
Same feature set was also used in various other experiments such as the INTERSPEECH Paralinguistic Challenges (2010-2016) \cite{schuller2010interspeech,schuller2011interspeech}. 
Section 3 in \cite{schuller2010interspeech} provides a complete description of these features.
We would like to note that the feature dimensionality is relatively high which renders the analysis of the data-points in the feature space challenging. 

\subsection{Experimental setup}
We conduct our experiments using a five fold cross-validation using one IEMOCAP session as a test set.
This ensures that that the models are trained and tested on speaker independent sets.
We initially provide a description of the adversarial auto-encoder training and then follow up with the two investigatory experiments regarding feature compression and synthetic data creation.

\subsubsection{Training the adversarial auto-encoder}
We train the adversarial auto-encoder on the training partition consisting of 4 sessions.
We use the adversarial auto-encoder setup that incorporates the label information in adversarial regularization, as described in Section 2.3 in \cite{makhzani2015adversarial}.
We chose the arbitrary distribution ($p(z)$ in \cite{makhzani2015adversarial}) to be a 4 component GMM in a $K$-dimensional subspace, to encourage each component to correspond to one of the four emotion labels. 
Our model is trained while the following two adversarial losses converge: (i) cross-entropy is minimized for code vectors to be classified as a synthetic sample (implying encoder is able to generate code vectors resembling the synthetic distribution), and (ii) cross-entropy is maximized for real versus synthetic data classification by the discriminator (discriminator maximally confuses between real and synthetic data). 
We summarize the training algorithm for the adversarial auto-encoder below, enlisting the specific parameter choices for our experiments.\\

\noindent While adversarial losses converge:

\begin{itemize}
\item Weights of the generator auto-encoder are updated based on a reconstruction loss function. We chose this function to be Mean Squared Error (MSE) between the inputs $x$ and the reconstruction $x^\prime$. Our encoder and decoder layers contain two hidden layers with 1000 neurons each. The auto-encoder is regularized using a dropout value of 0.5 for connection between every layer.

\item The data is transformed by the encoder and we sample an equal number of noise samples from the arbitrary PDF $p(z)$. Weights of encoder (in the generator's auto-encoder) and the discriminator are updated to minimize cross-entropy between real versus synthetic data labels. The discriminator model also consists of two hidden layers with 1000 neurons each.

\item We then freeze the discriminator weights. 
The weights of encoder are updated based on its ability to fool the discriminator (equivalently minimizing the cross-entropy for real samples to be labeled as synthetic).
\end{itemize}

\begin{figure}[t]
\begin{center}
\begin{tabular}
{@{\hspace{-0.0cm}}c@{\hspace{0.1cm}}c@{\hspace{0.0cm}}}
\raisebox{-.5\totalheight}{  \includegraphics[scale=0.25]{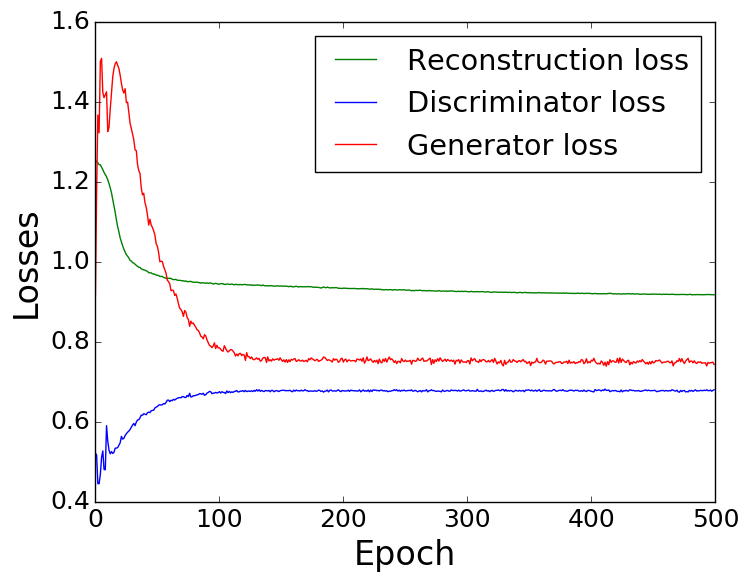}} \\
\raisebox{-.495\totalheight}{  \includegraphics[scale=0.25]{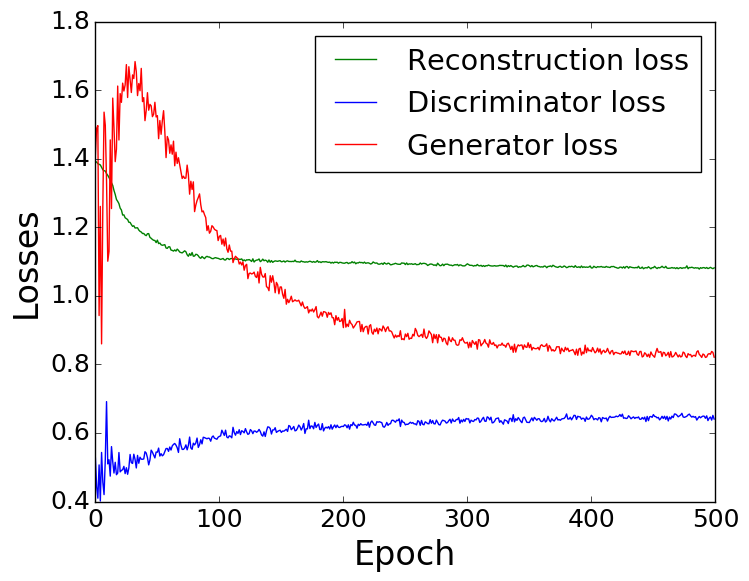}}\\
\end{tabular}
\end{center}
\vspace{-2mm}
\captionsetup{justification=centering}
\caption{Reconstruction and adversarial losses on the training set (top) and test set (bottom) for the adversarial auto-encoder. An increase in discriminator cross entropy loss indicates that the discriminator confuses more between real and synthetic samples and a decrease in generator cross entropy loss implies that more real samples are marked as synthetic. }
\vspace{-4mm}
\label{fig:errors}
\end{figure}

\begin{figure}[t]
\begin{center}
\begin{tabular}
{@{\hspace{-0.0cm}}c@{\hspace{0.1cm}}c@{\hspace{0.0cm}}}
\raisebox{-.5\totalheight}{\includegraphics[scale=0.25]{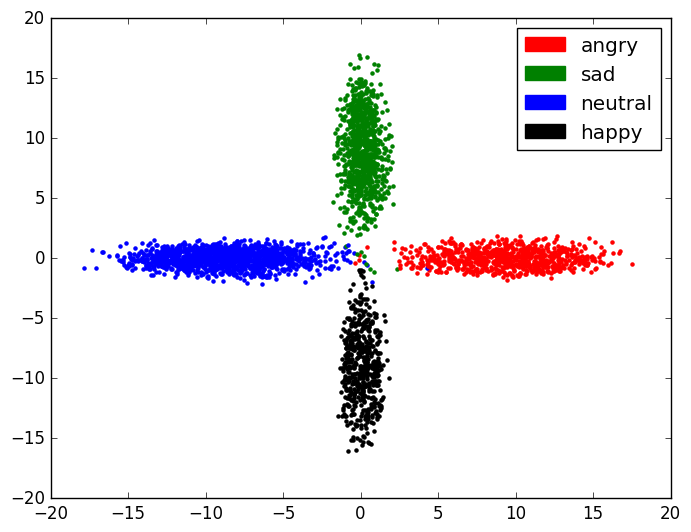}} \\
\raisebox{-.495\totalheight}{\includegraphics[scale=0.25]{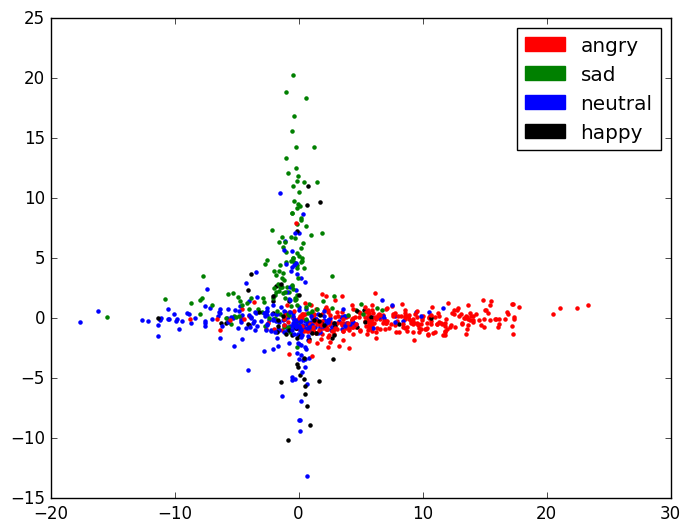}}\\
\end{tabular}
\end{center}
\vspace{-6mm}
\captionsetup{justification=centering}
\caption{Code vectors learnt on the 2-D encoding space for a specific partition for training set (top) and testing set (bottom) during the cross-validation.}
\vspace{-4mm}
\label{fig:2d-codes}
\end{figure}

We tune $K$ based on inner-fold cross validation on the training set, yielding $K=2$.
Upon increasing the encoder dimension, there was a minor decrease in accuracy which may be due to a greater overlap between the encoded vectors due to larger dimensionality. 
In Figure~\ref{fig:errors}, we look at the three metrics during adversarial auto-encoder training: the reconstruction error (MSE between $x$ and $x^\prime$), and the two adversarial cross-entropy losses. 
We plot these errors per epoch on the training and the testing set during one specific cross-validation set during the adversarial auto-encoder optimization.
We observe that while the reconstruction error decreases, the adversarial losses converge indicating that the discriminator's ability to discriminate is countered by generator's ability to confuse it.
This trend is observed for both, training and testing sets, indicating that the learnt parameters generalize well to data unseen during model training. 
After we train the adversarial auto-encoder, we use the auto-encoder in the generator to compute the code vectors for the training set as well as the testing set. 
Figure~\ref{fig:2d-codes} shows an example of the code vectors for the training and testing sets for one specific iteration during the cross-validation.
From the figure, we observe that while the training set instances from different classes are perfectly encoded into a specific component of the 4 component GMM model, the test set samples are also fairly separable. 
The figure provides a sense of the separability of emotion labels based on the 2-dimensional encodings of the 1582-dimensional openSMILE features.
The classification experiments quantify this separability. 

\subsubsection{Classification using the code vectors}
In this experiment, we quantify the discriminative ability of code vector and compare it against the full set of openSMILE features as well as a few other dimension reduction techniques.
The goal of this experiment is to quantify the loss in discriminability after compressing the original feature to a smaller feature subspace. 
For a specific cross-validation iteration, we train an SVM classifier on the openSMILE features as well as a lower dimensional representation of these features as obtained using the following techniques: Principal Component Analysis (PCA), Linear Discriminant Analysis (LDA), an auto-encoder and finally the code vector representations learnt using the adversarial auto-encoders. 
PCA, LDA and autoencoders have been investigated as dimensionality reduction techniques in similar experiments \cite{you2006emotion, cibau2013speech}. 
We learn and obtain these lower dimensional representations (PCA, LDA, auto-encoder and adversarial auto-encoder) of the openSMILE features on the training set, which are then used to train the SVM model.
The chosen projection dimensionality for PCA, LDA and the auto-encoder and the SVM parameters (box-constraint and kernel) are tuned using an inner-cross validation on the training set. Since our goal here is dimension reduction, we keep the maximum dimension of these representations during tuning to be 100 (note that setting projected PCA, auto-encoder dimensions to 1592 is equivalent to using the entire set of openSMILE features).
We use Unweighted Average Recall (UAR) as our evaluation metric as has also been done in previous works on the IEMOCAP dataset \cite{bone2014robust}.
We list the results of the classification experiment in Table~\ref{tab:UARs}.

\begin{table}[t]
\centering
\captionsetup{justification=centering}
\caption {Classification results on the openSMILE features, code vectors and the two feature sets combined.}
\begin{tabular}{@{}c|c|c|c|c|c} \hline
           & OpenSmile  & Code  & Auto-  & LDA & PCA \\ 
          & features  & vectors  & encoder & & \\
          & (1582-D) & (2-D) & (100-D) & (2-D) & (2-D) \\\hline
UAR (\%)      &57.88& 56.38 &  53.92   & 48.67 &  43.12  \\
\end{tabular}
\vspace{-6mm}
\label{tab:UARs}
\end{table}

From the results, we observe that the performances of SVMs trained on the openSMILE features and the code vectors are fairly close (the binomial proportions test for statistical significance yields a p-value=0.15) 
This indicates that the compressed code vectors capture the differences between the emotional labels in the openSMILE feature space to a fairly high degree.
We do not observe as high a performance from any of the other feature compression techniques. 
We also note that a vanilla auto-encoder does not perform as well as the adversarial auto-encoder, showing the value of class label based adversarial regularization. 
This low dimensional representation retaining the discriminability across classes provides a powerful tool for analysis in a low dimensional subspace, which is otherwise not possible with a large feature dimensionality. 
The low dimensional representation could be used for applications such as clustering as well as an ``experimentation by observation", as a low dimensional code vector (in particular 2-D) allows plotting the emotion utterances and analyzing them\footnote{For instance investigating the membership of test utterances in various GMM components based on Figure~\ref{fig:2d-codes} (bottom). We aim to conduct such an analysis in the future due to the required detail and maintaining the consistency of idea in this paper.}. 
We also note the fact that the auto-encoder allows reconstructing the features from these code vectors.
Therefore, a recovery of the actual utterance representations is also possible, which is otherwise more lossy in other dimension reduction techniques. 

\vspace{-2mm}
\subsubsection{Classification using synthetically generated samples}
\vspace{-1mm}
We next examine the possibility of synthetically creating samples representative of utterances with emotions. 
We randomly sample code vectors from each component of the GMM PDF.
The sampled code vector is then passed through the decoder part of the generator auto-encoder to yield a 1582 dimensional vector.
This synthetically generated vector thus is an openSMILE-like feature vector obtained by passing a randomly sampled 2-dimensional code vector through the decoder (and not directly obtained from an utterance from the database).
Note that each GMM component was enforced to pertain to a specific emotion label using the discriminator regularization.
The labels for the synthetically generated samples is assigned to be the same as the GMM component label used to sample the code vector.
In order to validate if the synthetic vectors have a correspondence to the vectors in the real dataset, we conduct another classification experiment with training on the synthetic dataset.
We initially train an adversarial auto-encoder on the training partition of the dataset.
We then sample 100 code vectors from each GMM component and generate a synthetic training set.
Next, we train an SVM to classify the test set under two settings: (i) using the synthetic dataset only and, (ii) appending the synthetic dataset to the available features from the real dataset.
Table~\ref{tab:UARs_syn} show the experimental results for this section.

\begin{table}[t]
\centering
\captionsetup{justification=centering}
\caption {Classification results on the openSMILE features, code vectors and the two feature sets combined.}
\begin{tabular}{c|c} \hline
Dataset  & UAR (\%)  \\ \hline
Chance accuracy & 25.00 \\
Synthetic datapoints only & 33.75 \\
Real datapoints only& 57.88\\
Synthetic + real datapoints & {\bf 58.38}\\ 
\end{tabular}
\vspace{-6mm}
\label{tab:UARs_syn}
\end{table}

From the results, we observe that a model trained on only synthetic dataset performs significantly above the chance model (binomial proportions test, p-value$<$0.05).
This indicates that the model trained solely on synthetic features does carry some discriminative information to classify utterances from the real dataset.
Addition of this synthetic dataset to the original dataset does marginally (although not significantly) increase the overall UAR performance. 
This encourages us to further investigate adversarial auto-encoders for synthetic data generation.
We note that the generated features do not follow the actual marginal distribution (marginalized over the class distribution) of the samples in the real dataset, as the marginal distribution is determined by the random sampling strategy for the code vectors.
We aim to address this issue in a future study.

\section{Conclusion}
\label{sec:conclusion}
\vspace{-2mm}

Automatic emotion recognition is a problem of wide interest with implications on understanding human behavior and interaction. 
A typical emotion recognition system design involves use of high dimensional features on a curated dataset.
This approach suffers from the drawbacks of limited dataset and challenging analysis in the high dimensional feature space. 
We addressed these issues using the adversarial auto-encoder framework.
We establish that the code vectors learnt by the adversarial auto-encoder can be obtained in a low dimensional subspace without losing the class discriminability in the higher dimensional feature space.
We also observe that synthetically generating samples from an adversarial auto-encoder shows promise as a method for improving the classification of data from the real world.

Future investigations include a detailed analysis of emotional utterances in the low dimensional code vector space. 
We aim to investigate and further improve the classification schemes using synthetic vectors. Additionally we plan to investigate auto-encoder architectures that can be fed frame level features instead of utterance level features. We believe temporal dynamics of feature contours can lead to better classification results.
Finally, the adversarial auto-encoder architecture can also be used in analysis of other behavioral traits such as engagement \cite{gupta2016analysis} jointly with the emotional states.

\newpage
\bibliographystyle{IEEEbib}
\bibliography{strings}
\end{document}